\documentclass[10pt,twocolumn,letterpaper]{article}

\usepackage{cvpr}              % To produce the CAMERA-READY version
\usepackage{graphicx}
\usepackage{amsmath}
\usepackage{amssymb}
\usepackage{booktabs}
\usepackage{bbding}

\usepackage[pagebackref,breaklinks,colorlinks]{hyperref}

\usepackage{adjustbox}

\usepackage[capitalize]{cleveref}
\crefname{section}{Sec.}{Secs.}
\Crefname{section}{Section}{Sections}
\Crefname{table}{Table}{Tables}
\crefname{table}{Tab.}{Tabs.}

\def\confName{CVPR}
\def\confYear{2023}

\begin{document}

%%%%%%%%% TITLE - PLEASE UPDATE
\title{FLIGHT Mode On: A Feather-Light Network for Low-Light Image Enhancement}

\author{Mustafa Özcan, Hamza Ergezer, Mustafa Ayazaoğlu \\
Aselsan Research\\
Ankara, Turkey\\
{\tt\small \{mustafaozcan, hergezer, mayazoglu\}@aselsan.com.tr}
% For a paper whose authors are all at the same institution,
% omit the following lines up until the closing ``}''.
% Additional authors and addresses can be added with ``\and'',
% just like the second author.
% To save space, use either the email address or home page, not both
%\and
%Hamza Ergezer\\
%Aselsan Research\\
%Ankara, Turkey\\
%{\tt\small hergezer@aselsan.com.tr}
%\and
%Mustafa Ayazoglu\\
%Aselsan Research\\
%Ankara, Turkey\\
%{\tt\small mayazoglu@aselsan.com.tr}
}
\maketitle

%%%%%%%%% ABSTRACT
\begin{abstract}
Low-light image enhancement (LLIE) is an ill-posed inverse problem due to the lack of knowledge of the desired image which is obtained under ideal illumination conditions. Low-light conditions give rise to two main issues: a suppressed image histogram and inconsistent relative color distributions with low signal-to-noise ratio. In order to address these problems, we propose a novel approach named FLIGHT-Net using a sequence of neural architecture blocks. The first block regulates illumination conditions through pixel-wise scene dependent illumination adjustment. The output image is produced in the output of the second block, which includes channel attention and denoising sub-blocks. Our highly efficient neural network architecture delivers state-of-the-art performance with only 25K parameters. The method's code, pretrained models and resulting images will be publicly available.
\end{abstract}

%%%%%%%%% BODY TEXT
\section{Introduction}
\label{sec:intro}

A low-light image can be defined as an image captured in deficient illumination conditions that do not fully excite the detector. Hence, the output image is not even close to have ideal histogram distributions. In such conditions, a dedicated algorithm enhancing the image is needed to present a better image and to to help increasing the performance of consequent blocks which are trained or tuned under normal lightning conditions, such as object detection, etc.

Global or local image histogram equalization techniques \cite{dipbook, clahe} are the first-thought candidates to solve the low-light image enhancement problem. However, they do not employ spatial information and work in pixel level that does not include surrounding content information. On the other hand, using deep neural network architectures, spatial information can be utilized and combined with color information in different scales. Therefore, deep neural architectures recently provide superior performance for low-light image enhancement problem as in most of other low-level and high-level vision problems. 
%%%%Figure
\begin{figure}[t]
\begin{center}
% \fbox{\rule{0pt}{2in} \rule{0.9\linewidth}{0pt}}
% 
\includegraphics[scale=0.50]{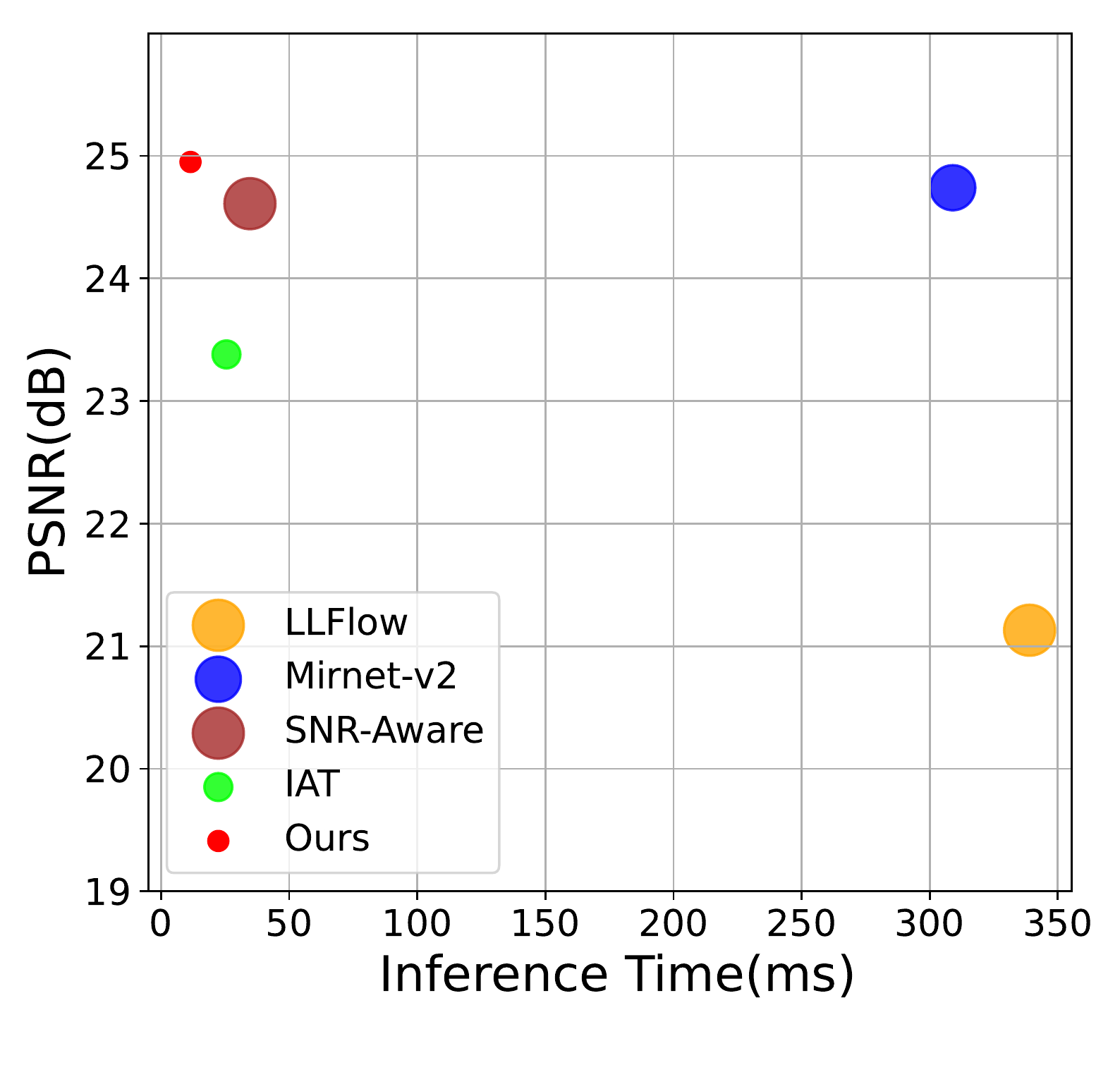}
\vspace{-15pt}
\caption{Performance Comparison on LOL-v1 Dataset. The diameters of the circles are proportional to the number of model parameters.}
\vspace{-20pt}
\label{fig:long}
\label{fig:onecol}
\end{center}
\end{figure}

%%%%Figure-end

For most of the applications, low-light image enhancement should be implemented in image signal processing(ISP) framework of a camera. ISP frameworks process the data in real-time, and include some certain blocks. Hence, it can be argued that the most crucial requirements for LLIE should be computational memory and complexity while not sacrificing from the visual quality.  In recent years, there are some studies \cite{cycleISP, ispeccv22eth} that handles all the blocks of ISP blocks in a single network. However, it is not easy to deploy such a network of multi-million parameters in an edge device, i.e. surveillance camera. Therefore, it is crucial to have lightweight blocks to handle unusual conditions in ISP framework. The proposed solution is a good candidate for such cases to deploy in an edge device.

Inspired by several recent studies \cite{90K, cycleISP}, we propose a feather-light network with carefully designed blocks to match the problem's nature at the hand instead of throwing all of the information into a huge network and hope to get the right output by some "deep magic". As with the previous studies, we model the LLI as an image generated as reflectance multiplied by illumination. For this purpose, first, we strive to achieve pixelwise scene illumination. Using such an approach, we achieve an equalized the image histogram while handling uneven illumination at the same time. Since the input signal has very low SNR, there exist inherent noise in the input and this becomes visible with the illumination adjustment, the noise and color inconsistencies should be taken care of by the subsequent blocks. To solve these problems, we utilize the ideas presented in recent approaches \cite{cycleISP, uw1, uw2} that addresses these problems using channel and spatial attention blocks \cite{channelatt, cbam}. In simple terms these attention mechanisms enable selective feature extraction for image enhancement and denoising.

We propose a novel efficient neural network architecture named FLIGHT-Net is proposed for low-light image enhancement problem. It is shown that FLIGHT-Net gives outstanding results compared the state of the art. To the best of our knowledge, it is the lightest network that achieves great balance between run-time and performance among supervised learning LLIE methods.

\section{Related Work}

As in high-level vision tasks, the solutions based on deep neural architectures provide most successful results for image restoration tasks. Before deep learning era, traditional methods also give satisfactory results up to a point. Global and local histogram equalization methods \cite{dipbook, clahe} are most well-known solutions for the LLIE problem. Furthermore, Retinex theory increased the understanding of the problem from a more theoretical point of view. Following Retinex theory, in LIME \cite{lime}, the overall solution is based on illumination map estimation (IME) approach. Although the IME approach is inspiring, being a traditional method LIME \cite{lime} failed to generalize well on different scenarios. On the other hand, IME block kept its existence on different deep learning approaches such as \cite{sparseTIP2021,90K} and it is also a part of our proposed solution.   

Starting from the pioneer work \cite{llnet}, the deep learning based LLIE methods can be categorized according to general deep learning strategies. In other words, deep LLIE methods can be divided into four main categories as supervised learning \cite{snraware, 90K, llflow, retinexnet, kind}, semi-supervised learning, zero-shot learning \cite{zerodce,zerodcepp, retinexdip} and unsupervised learning. Although the number of other types of methods is also notable, the core part of the literature is formed by supervised and zero-shot learning based methods. 

In supervised learning based methods, there are two main approaches. The first approach is to extract the enhanced image by using a single end-to-end network \cite{dslr,llflow}. The second and mostly utilized approach is to design the subnetworks according to the Retinex theory \cite{retinex}. In these approaches, subnetworks are designed to reconstruct the illumination and refleftance parts of the enhanced image. In \cite{retinexnet}, two main blocks called Decom-Net and Enhance-Net are used to extract illumination and reflectance maps and then adjust the illumination according to the decomposed maps. Illumination adjustment is handled by pixelwise enhancement block in \cite{90K}, while color correction is solved using a transformer block.

It is not easy to build a setup with the ground truth and low-light image pair required by the problem. Therefore, more recently, zero-shot learning methods \cite{zerodce, zerodcepp, retinexdip, ruas} are proposed for LLIE problem. Zero-DCE and its extended version \cite{zerodce, zerodcepp} solve the problem by predicting a set of high-order curves for a given image. In \cite{retinexdip}, a generative strategy is applied to decompose the illumination and reflectance components. After the decomposition, enhanced image is obtained by processing the illumination component. RUAS \cite{ruas} utilizes again Retinex theory and neural architecture search strategy to determine the basic blocks called illumination estimation and noise removal module. Low-light enhancement problem is also elaborated as a sub-problem in some recent image restoration studies \cite{mirnetv1, mirnetv2, maxim}.

%%%%%%%FIGURE%%%%%%
\begin{figure*}[ht!]
	\centering
	\begin{subfigure}{1.0\textwidth}
		\includegraphics[width=\textwidth]{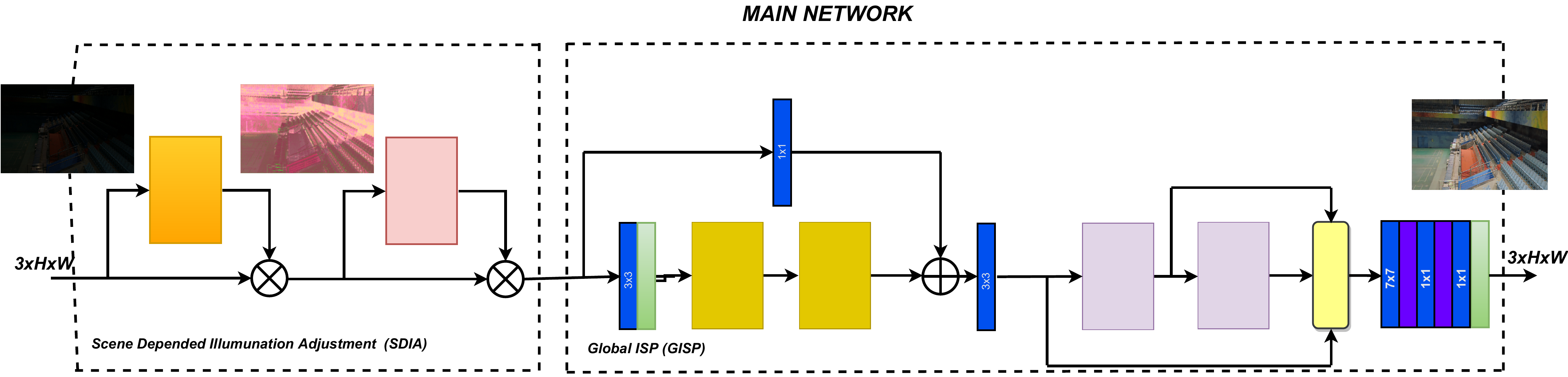}
		\caption{Main Network}
        \label{fig:Network_main_a}
	\end{subfigure}
 %%%%%%%%%%%%%%
	\begin{subfigure}{0.9\textwidth}
		\includegraphics[width=\textwidth]{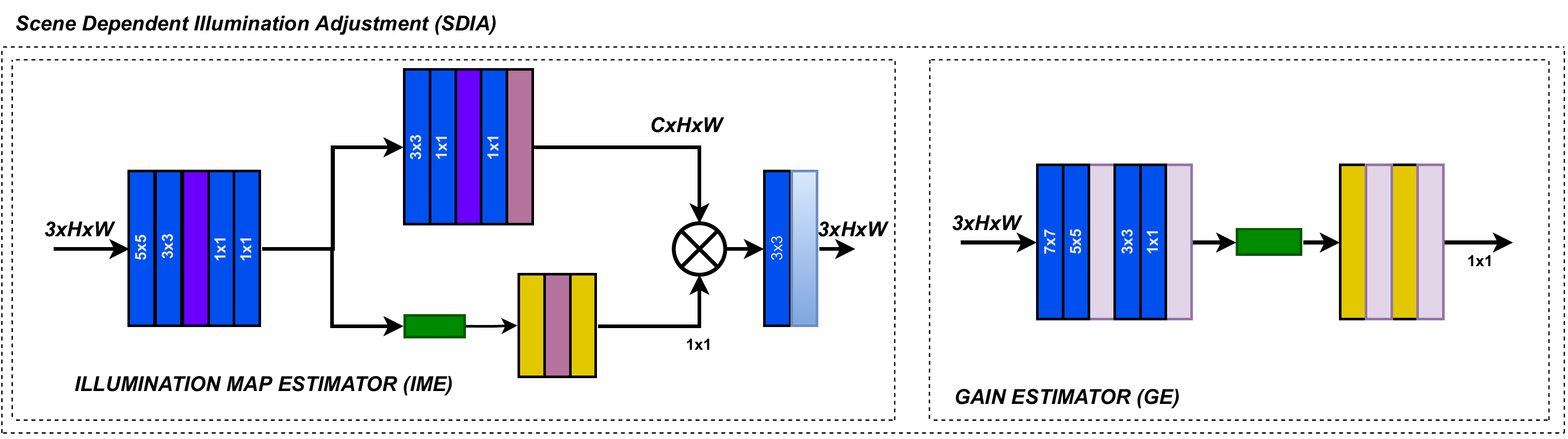}
		\caption{SDIA Block}
        \label{fig:Network_main_SDIA}
	\end{subfigure}
 %%%%%%%%%%%%%%
	\begin{subfigure}{0.9\textwidth}
		\includegraphics[width=\textwidth]{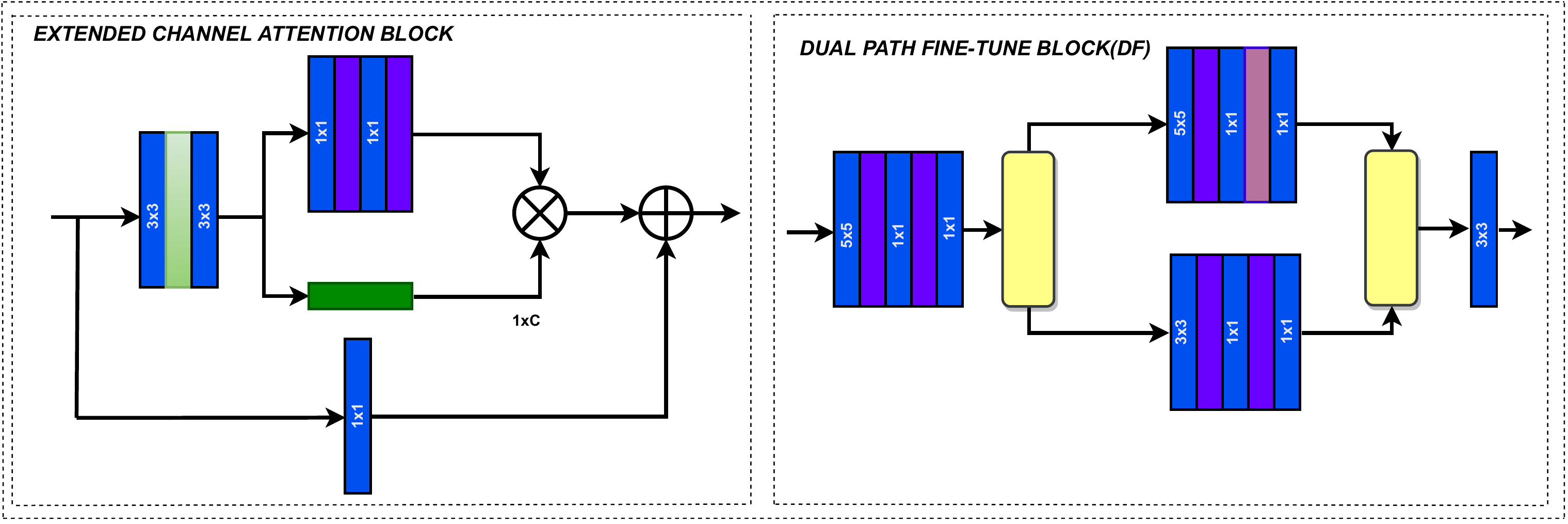}
		\caption{GISP Block}
        \label{fig:Network_main_GISP}
	\end{subfigure}
  %%%%%%%%%%%%%%
	\begin{subfigure}{0.8\textwidth}
		\includegraphics[width=\textwidth]{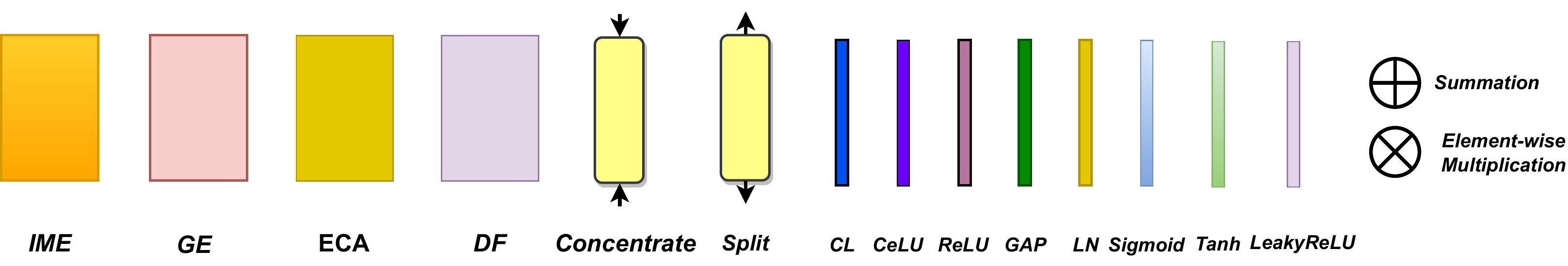}
		\caption{IME:Illumination Map Estimation, GE:Gain Estimator, ECA:Extended Channel Attention, DF:Dual Path Fine-Tune, CL: Convolutional Layer, CeLU: Continuously Differentiable Exponential Linear Unit, ReLU: Rectified Linear Unit, GAP: Global Average Pooling, LN: Linear Layer}
        \label{fig:Network_main_item_details}
	\end{subfigure}
  %%%%%%%%%%%%%%
	\caption{FLIGHT-NET}
    \label{fig:Network_main}
    \vspace{-8pt}
\end{figure*}

%%%%%%%%%%%%%%%%%%%%%%%

\section{Proposed Method}

LLIE can be posed as a reconstruction of the ideal image which is under the ideal light conditions.
In this section, the motivation and formulation of FLIGHT-Net are introduced. Then the details of the network are presented.

\subsection{Problem Formulation}
The proposed method and formulation are inspired by Retinex theory\cite{retinex} and ISP framework\cite{ignatov2020replacing}. As it is known, in the Retinex theory, an image consists of reflection and illumination via equation \ref{eq:1}: 

\vspace{-8pt}
\begin{align}
\label{eq:1}
I = R.L
\end{align}
where $I,R,L$ donate image, reflectance and illumination. Retinex-based deep learning methods estimate an illumination map using low-light image \cite{retinexdip}. Estimated illumination maps which is required to obtain a normal light image from a low light image, transform the image on a pixel-by-pixel basis. Furthermore, considering the image acquisition from the camera, a series of transformations are applied to the linear raw RGB image, which makes obtaining normal-light image from low-light image more challenging. The transformation can be expressed as in equation \ref{eq:2}. 

\begin{align}
\label{eq:2}
I=f_I{_S{_P}} (S)
\end{align}
where S denotes camera input and $f_I{_S{_P}}$ denotes the whole transformation function obtaining sRGB image from sensor input, i.e. raw data. The quality of an low-light image is affected by two main factors: low illumination and in-camera noise. The sensor input in the sensor layer produces a Raw linear RGB image that is linearly related to the ambient light. To obtain an image similar to a standard light image, the raw linear RGB image can be enhanced using an appropriate gain ratio, eliminating the first degradation of the low-light image. However, the second degradation, in-camera noise, increases linearly with the gain ratio and distorts the image.

Another crucial issue when working with low-light sRGB images is that raw image is processed through a series of non-linear operations such as white balance, gamma correction, noise reduction, contrast enhancement, and edge enhancement, hence it is no longer possible to obtain an normal-light sRGB image with the appropriate gain coefficient using the low-light sRGB image. Cui et.al \cite{90K} explains this by stating that the actual luminance degradation occurs in the raw-RGB space in the ISP framework and proposed the IAT network design, which is inspired and characterized the ISP process.
 
Inspired by Retinex theory\cite{retinex}, the ISP framework \cite{ignatov2020replacing} and IAT network \cite{90K}, a new deep learning network called FLIGHT-Net for low-light image enhancement is proposed. FLIGHT-Net is comprised of two primary network blocks: the Scene Dependent Illumination Adjustment (SDIA) block, and the Global ISP block (GISP). The SDIA block modifies the input in a pixel-wise manner, while the GISP block transforms the image globally. The formulation of the proposed method is shown in equation \ref{eq:3}:
\begin{align}
\label{eq:3}
NLI=f_G{_I}{_S}{_P}(f_S{_D}{_I}{_A}(LLI))
\end{align}
where $LLI, NLI, f_S{_D}{_I}{_A},f_G{_I}{_S}{_P}$ represent low-light image, normal-light image, SDIA and GISP blocks respectively. 

SDIA block is composed of two separate blocks:  illumination map estimator and gain estimator. As stated earlier, a low-light image is an image in which the sensor receives less light than a normal-light image. The gain estimator predicts the required gain ratio necessary to facilitate image enhancement. However, an accurate gain ratio for LLIE is not possible due to the ISP applied when capturing the sRGB image. Therefore, the image is transformed using a suitable map (IM), which is the output of the  illumination map estimator (IME) block to increase the effect of the gain value. As a result, the input image is mapped to the latent feature space by utilizing the SDIA. 

During the process of obtaining the normal-light image from the low-light image, in the SDIA, noise becomes more apparent. This is due to the inherent noise in the low-light image due to low PSNR and when low-light is illumination adjusted the noise is also becomes visible and deteriorates the visual quality. However, since the noise is visible along with the scene itself another network might be selectively diminish the noise while preserving or even enhancing the scene. Indeed our GISP block suppresses the noise while keeping structures intact through embodied attention mechanisms as shown \ref{fig:Network_main_GISP}. We can consider GISP block as a transformation from a latent feature space to sRGB space. In other words, color correction, denoise and white balance operations of a typical ISP framework are mimicked via GISP block.

\subsection{Network Framework}

FLIGHT-Net consists of two main network block as shown in Figure \ref{fig:Network_main_a}. The first block that converts  input locally is Scene Depended Illumination Adjustment (SDIA) Network Block, and other block Global ISP Network Block (GISP) that transform its input globally. SDIA include IME and GE block whose outputs are multiplied by the input image on a pixel-wise basis. On the other hand, the output of the GISP block is used directly as the network output. 

As previously stated, the low-light image is characterized by insufficient light exposure onto the sensor, resulting in a dark and low dynamic range projection. Multiplying low-light image with a single gain coefficient can be considered as a naive way for obtaining a more pleasing image. However, this is usually not enough due to the non-linear effects introduced in the ISP framework such as local histogram enhancement where each pixel are subject to different illumination and hence need to be individually corrected, so, simply adjusting the gain coefficient alone is not sufficient. Since it may cause over-illumination in some parts, while leaving other areas in darkness. To address this issue more effectively, an IME block has been devised to estimate the illumination adjustment coefficients required for each pixel. Through this estimation process, the input image can be better prepared for gain adjustment block, ultimately leading to a more efficient utilization of the gain coefficient.

The IME block, illustrated in Figure \ref{fig:Network_main_SDIA}, includes CNN blocks for feature extraction subblocks and local gain coefficient. After feature extraction, the illumination map can be estimated effectively by multiplying map features and the local gain coefficient estimated with the help of LN by the features. At the end of the IME block, the sigmoid activation function is preferred because the IME block is designed to transform the image where the gain coefficient works efficiently. 
%If the IME block ends with any activation function whose output can be greater than one, the IME block also takes over the job of improving visibility. In this case, network performance is adversely affected. %This is also an undesirable situation. because if each block fulfills its intended purpose, the best results will be obtained by using the least model parameters.

\begin{table*}[h]

\centering
\begin{adjustbox}{width=1\textwidth}
\begin{tabular}{l|llllllll}

\hline
Method & \multicolumn{1}{c}{SNR-ALLIE\cite{snraware}}& RetiNexNet\cite{Retinex-Net_wei2018deep}&MBLLEN\cite{mbllen_lv2018}& DRBN\cite{DRBN}&KIND\cite{kind}&MAXIM\cite{maxim}&IAT\cite{90K}& \textbf{Ours}\\
\hline
\hline
PSNR   & 24.61      & 16.77      & 17.90  & 19.55 & 20.86 & \textit{23.43} & 23.38 & \textbf{24.96} \\
SSIM   & 0.842      & 0.562      & 0.70  & 0.746 & 0.790 & \textbf{0.863} & 0.809 & \textit{0.85}  \\ 
\hline

\end{tabular}
\end{adjustbox}
\vspace{-10pt}
\caption{Comparative performance results for LOL-v1 dataset}
\label{tab:lolv1}
\vspace{+8pt}
% \end{table*}

%%%%%%%%%%%%%%%%%%%%%%%%%%%%%%%%%%%%%%%%%

% \begin{table*}[htp2]
\centering
\begin{adjustbox}{width=1\textwidth}
\begin{tabular}{l|llllllll}

\hline
Method & \multicolumn{1}{c}{SNR-ALLIE\cite{snraware}} & Retinex\cite{ruas} & IPT\cite{IPT_chen2021pre}   & Sparse\cite{sparseTIP2021} & Band\cite{band_yang2021}  & MIR-Net\cite{mirnetv1} & LPNet\cite{LP_Net_li2020luminance} & \textbf{Ours}      \\
\hline
\hline
PSNR   & 21.48                         & 18.37   & 19.80 & 20.06  & \textit{20.29} & 20.02   & 17.80 & \textbf{21.71} \\
SSIM   & 0.849                         & 0.723   & 0.813 & 0.816  & \textit{0.831} & 0.820   & 0.792 & \textbf{0.834} \\ \hline
\end{tabular}
\end{adjustbox}
\vspace{-10pt}
\caption{Comparative performance results for LOL-v2-Real dataset}
\label{tab:lolv2r}
\vspace{+8pt}

%%%%%%%%%%%%%%%%%%%%%%%%%%%%%%%%%%%%
% \begin{table*}[htp2]
\centering
\begin{adjustbox}{width=1.\textwidth}
\begin{tabular}{l|llllllll}

\hline
Method & \multicolumn{1}{c}{SNR-ALLIE\cite{snraware}} & Retinex\cite{ruas} & IPT\cite{IPT_chen2021pre}   & Sparse\cite{sparseTIP2021} & Band\cite{band_yang2021}  & MIR-Net\cite{mirnetv1} & LPNet\cite{LP_Net_li2020luminance} & \textbf{Ours}      \\
\hline
\hline
PSNR   & 24.14                         & 16.55   & 18.30 & 22.05  & \textit{23.22} & 21.94   & 19.51 & \textbf{24.92} \\
SSIM   & 0.928                         & 0.652   & 0.811 & 0.905  & \textit{0.927} & 0.876   & 0.846 & \textbf{0.93} \\ \hline
\end{tabular}
\end{adjustbox}
\vspace{-10pt}
\caption{Comparative performance results for LOL-v2-Synthetic dataset}
\label{tab:lolv2s}
\vspace{+8pt}

%%%%%%%%%%%%%%%%%%%%%%%%%%%%%%%%%%%%
% \begin{table*}[htp2]
\centering

\begin{adjustbox}{width=\textwidth}
\begin{tabular}{l|llllllll}
\hline
Method & \multicolumn{1}{c}{Zero-DCE\cite{zerodce}} & LIME\cite{lime} & Retinex-Net\cite{Retinex-Net_wei2018deep} & KinD\cite{kind}  &MBLLEN\cite{mbllen_lv2018} & GLADNet\cite{gladnet_wang2018} & MIR-Net\cite{mirnetv1} & \textbf{Ours}      \\
\hline
\hline
PSNR   & 12.99                         & 14.95   & 15.43 & 15.84 &17.52 & 21.09 & \textit{21.62} & \textbf{22.44} \\
SSIM   & 0.44                         & 0.45   & 0.34 & 0.49  & 0.60 & 0.69   & \textit{0.77} & \textbf{0.794} \\ \hline
\end{tabular}
\end{adjustbox}
\vspace{-10pt}
\caption{Comparative performance results for Rellisur dataset}
\label{tab:rellisur}
\end{table*}

The GE block in Figure \ref{fig:Network_main_SDIA} is used to estimate the appropriate gain coefficient depending on its input which might be captured varying light conditions. It consists of CNN blocks to extract features and a linear layer to estimate the required gain coefficient using these extracted features. After gain adjustment, the input image is converted to the latent feature, which is needed for GISP to deliver a normal-light sRGB image on its output.

The GISP block is the second main block of FLIGHT-Net. As stated previously, the main purpose of this block is color correction and denoising. It consists of extended channel attention (ECA) block and dual path fine tune (DF) blocks. The ECA block is the extended version of the CA block in \cite{CA_zhang2018image}. This block mainly strengthens the information like structures and patterns in the necessary channels while surpassing the unwanted information like noise. In the DF block, the extracted features in the channel are divided into two in order to transform features with different receptive field, different activation functions and kernel sizes in convolution operations. In addition, foremental information are carried forward and information at every stage is protected. The details of the GISP block are shown in Figure \ref{fig:Network_main_GISP}.

Last but not least we tried to reflect the current network design language to our design by using some of the suggestions for CNNs for reaching Transformer like performance as suggested in \cite{liu2022convnet}. It is reported that CNN architectures can obtain the success of the transformers with a proper selection of some parameters. For instance, as it is known in classical CNN design, convolution kernel size 3x3 is mostly preferred. In the FLIGHT-Net, 5x5 and 7x7 kernel sizes are also preferred in the convolution operation, especially at the beginning of the blocks used for feature extraction. In conclusion, it seems that the results in \cite{liu2022convnet} is indeed be beneficial for LLIE.
%%%
\begin{table}[htp]
\centering
%\begin{adjustbox}{width=\textwidth/2}

\begin{tabular}{|l|c|c|}

\hline
Dataset     & Training Image          & Validation Image                  \\ \hline
LOL\cite{Retinex-Net_wei2018deep}     & 485                & 15           \\
LOL-v2-Syn\cite{DRBN}                 & 900                & 100           \\
LOL-v2-Real\cite{DRBN}                & 689                & 100            \\
Rellisur\cite{rellisur_aakerberg2021}    & \begin{tabular}[c]{@{}c@{}}3610 LL\\ 722 NL\end{tabular} & \begin{tabular}[c]{@{}c@{}}215 LL\\ 43 NL\end{tabular} \\ \hline
\end{tabular}
%\end{adjustbox}
\vspace{-8pt}
\caption{Datasets used in experiments *low-light (LL), normal-light(NL)}
\label{table:Used_Datasets}
\end{table}
%%%

%%%
\begin{table}[htp]
\centering
\begin{tabular}{|l|l|l|l|l|}
\hline
Method & PSNR & SSIM  & \#P(M) & AT(ms) \\ 
\hline

Zero-DCE++ \cite{zerodcepp}                  & 14.83                    & 0.531 & 0.01            & 1.02   \\
LLFlow \cite{llflow}                    & 21.13                    & 0.853 & 38.86              & 339.69   \\
Mirnet-v2 \cite{mirnetv2}                        & 24.74                    & 0.851 & 5.86           & 309.68   \\
SNR-Aware \cite{snraware}                    & 24.61                    & 0.842 & 39.12           & 34.67   \\
IAT \cite{90K}                         & 23.38                    & 0.809 & 0.09            & 25.50   \\
Ours                         & 24.96                    & 0.85  & 0.025            & 11.47   \\ \hline
\end{tabular}

\caption{Comparative comparison of the methods according to the number of parameters (\#P) and average computation time (AT) on LOL-v1 dataset without using GT images.}
\label{tab:computation}

\end{table}

%%%%%
%%%%%%%%
\begin{figure}[ht!]
	\centering
	\begin{subfigure}{.2\textwidth}
		\includegraphics[width=\textwidth]{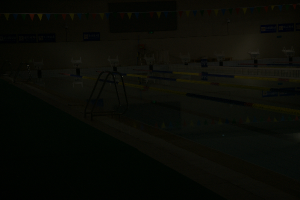}%\llap{\includegraphics[height=1.2cm]{CVPR2023-UG2-LLIE_v1/images/Figure_For_ablation_study/Crp_1_GT.png}}
		\caption{Input from LOLv2-Real}
	\end{subfigure}
 %%%%%%%%%%%%%%
	\begin{subfigure}{.2\textwidth}
		\includegraphics[width=\textwidth]{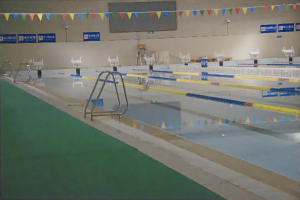}%\llap{\includegraphics[height=1.2cm]{CVPR2023-UG2-LLIE_v1/images/Figure_For_ablation_study/Crp_1_wog.png}}
		\caption{Estimated Image}
	\end{subfigure}
 %%%%%%%%%%%%%%
	\begin{subfigure}{.2\textwidth}
		\includegraphics[width=\textwidth]{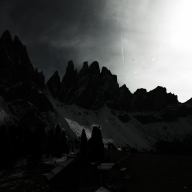}%\llap{\includegraphics[height=1.2cm]{CVPR2023-UG2-LLIE_v1/images/Figure_For_ablation_study/Crp_1_wod.png}}
		\caption{Input from LOLv2-Syn}
	\end{subfigure}
  %%%%%%%%%%%%%%
	\begin{subfigure}{.2\textwidth}
		\includegraphics[width=\textwidth]{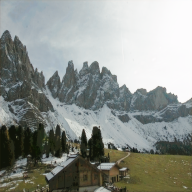}%\llap{\includegraphics[height=1.2cm]{CVPR2023-UG2-LLIE_v1/images/Figure_For_ablation_study/Crp_1_o.png}}
		\caption{Estimated Image}
	\end{subfigure}
  %%%%%%%%%%%%%%
  
  %%%%%%%%%%%%%%
  
	\begin{subfigure}{.2\textwidth}
		\includegraphics[width=\textwidth]{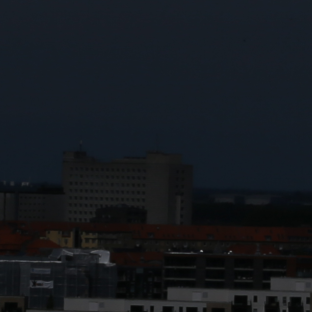}%\llap{\includegraphics[height=1.2cm]{CVPR2023-UG2-LLIE_v1/images/Figure_For_ablation_study/Crp_2_GT.png}}
		\caption{Input from Rellisur}
	\end{subfigure}
 %%%%%%%%%%%%%%
	\begin{subfigure}{.2\textwidth}
		\includegraphics[width=\textwidth]{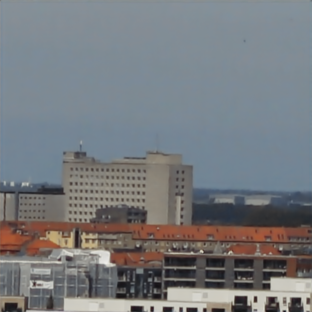}%\llap{\includegraphics[height=1.2cm]{CVPR2023-UG2-LLIE_v1/images/Figure_For_ablation_study/Crp_2_wog.png}}
		\caption{Estimated Image}
	\end{subfigure}
 %%%%%%%%%%%%%%
	\begin{subfigure}{.2\textwidth}
		\includegraphics[width=\textwidth]{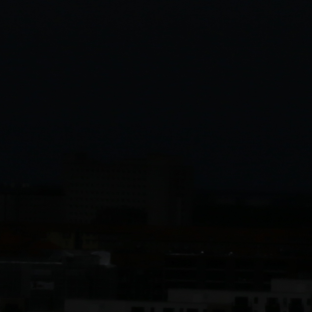}%\llap{\includegraphics[height=1.2cm]{CVPR2023-UG2-LLIE_v1/images/Figure_For_ablation_study/Crp_2_wod.png}}
		\caption{Input from Rellisur}
	\end{subfigure}
  %%%%%%%%%%%%%%
	\begin{subfigure}{.2\textwidth}
		\includegraphics[width=\textwidth]{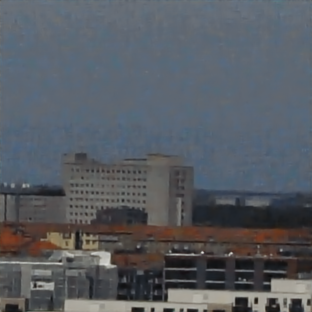}%\llap{\includegraphics[height=1.2cm]{CVPR2023-UG2-LLIE_v1/images/Figure_For_ablation_study/Crp_2_o.png}}
		\caption{Estimated Image}
	\end{subfigure}
  %%%%%%%%%%%%%%

	\caption{Visual results for LOL-v2-real, LOL-v2-synthetic and Rellisur datasets.}
\end{figure}
%%%%%

\section{Experiments}

%%%%Figure
\begin{figure}[htp]
\begin{center}
\begin{subfigure}{.45\textwidth}
% \fbox{\rule{0pt}{2in} \rule{0.9\linewidth}{0pt}}
\includegraphics[width=\textwidth]{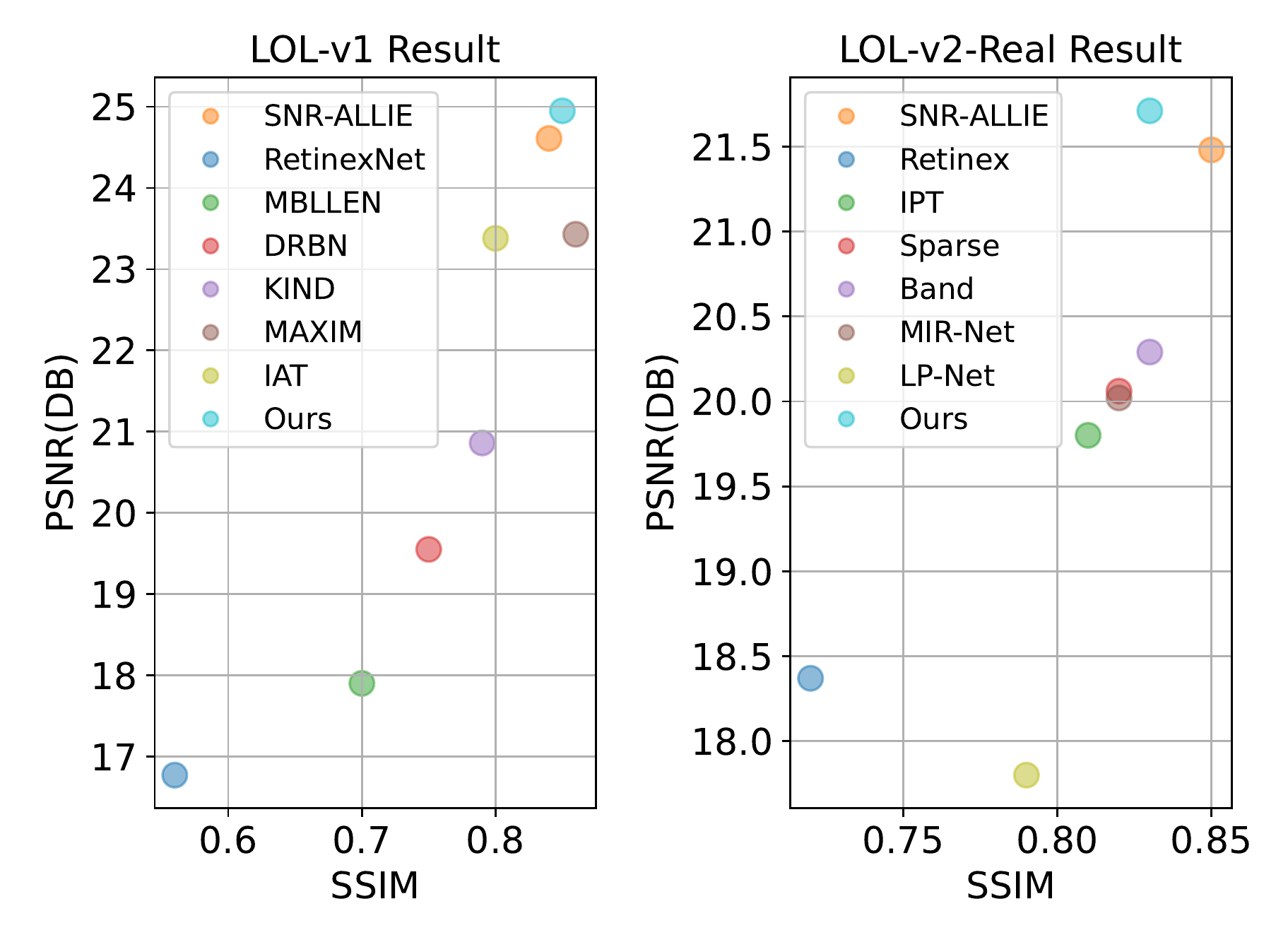}
\end{subfigure}

\begin{subfigure}{.45\textwidth}
% \fbox{\rule{0pt}{2in} \rule{0.9\linewidth}{0pt}}
\includegraphics[width=\textwidth]{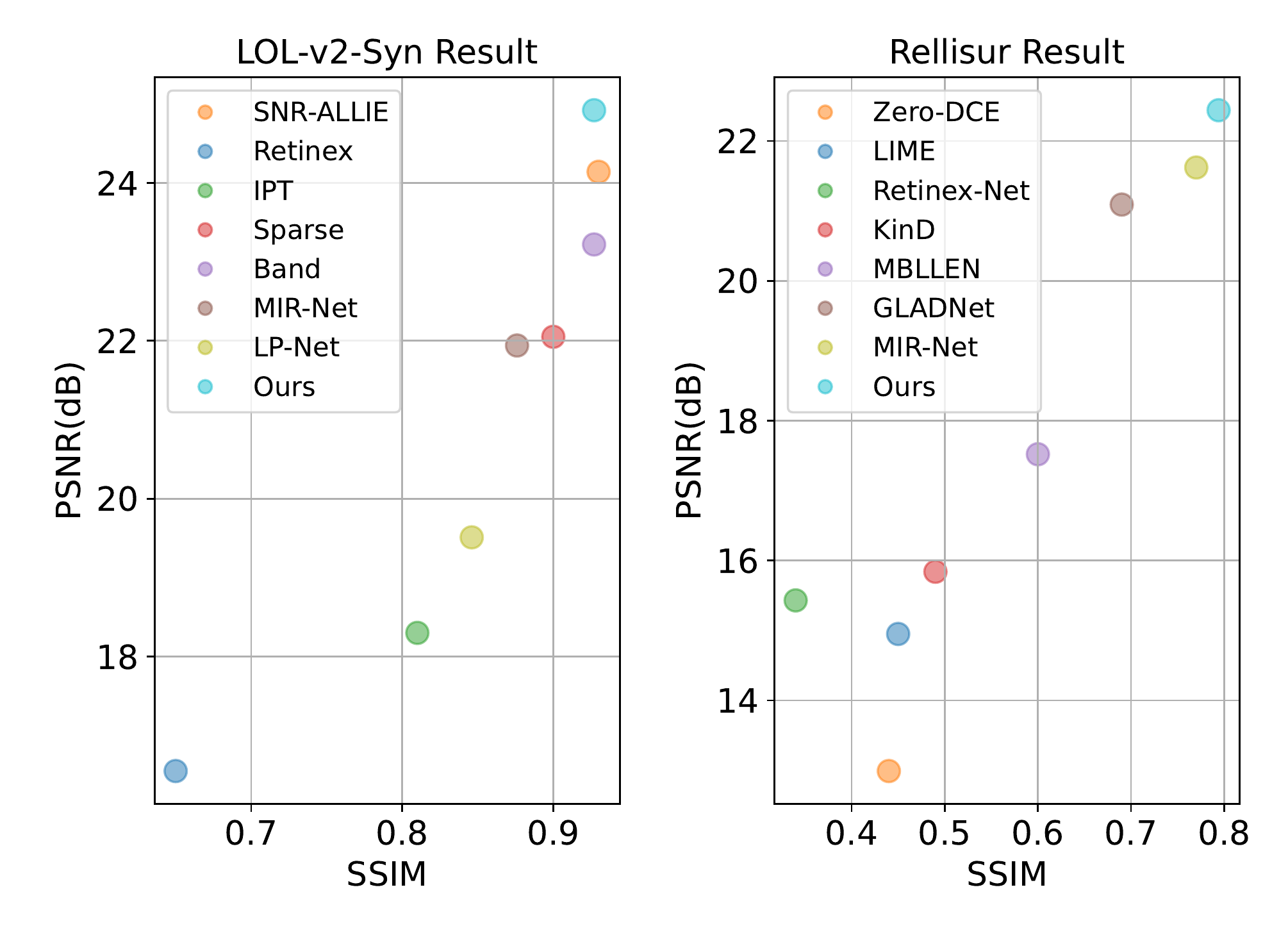}
\end{subfigure}
\caption{Performance Comparison}
\vspace{-20pt}
\label{fig:performance}
\end{center}
\end{figure}

%%%%Figure-end

%%%%%%%%%%%%
LOL-v1\cite{Retinex-Net_wei2018deep}, LOL-v2-real\cite{DRBN}, LOL-v2-synthetic\cite{DRBN} and Rellisur\cite{rellisur_aakerberg2021} datasets are chosen for our experiment. The LOL-v1 dataset is the first real dataset for low-light image enhancement and is used to test many state-of-art network\cite{90K, snraware, llflow, kind,retinexnet,DRBN}. The LOL-v2 dataset is the enhanced version of the LOL-v1 dataset. It contains two different training and validation image pairs, a real captured image and a synthetically acquired image. Finally, Rellisur dataset is the first multi-purpose dataset for the problems of low-light image enhancement and super-resolution. Details of the training and validation image pairs of the datasets used in the experiment are summarized in the Table \ref{table:Used_Datasets}.

During training, all experiments are performed in Tesla V100 GPU 16GB. ADAMW optimizer is selected in the training process as in \cite{liu2022convnet}. Batch size is set to 16 and initial leaning rate is $8\times10^{-4}$. The network is trained with initial learning rate for 6000 epoch and it reduced linearly 1/10th of it between 6000-12000 epoch. For fine tuning phase, previous scheduling is applied by starting from $4\times10^{-4}$.
%%%%%%%%%%%%

%%%%%%%%
\begin{figure*}[ht!]
	\centering
	\begin{subfigure}{.24\textwidth}
		\includegraphics[width=\textwidth]{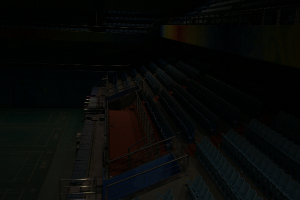}
		\caption{Input}
	\end{subfigure}
 %%%%%%%%%%%%%%
	\begin{subfigure}{.24\textwidth}
		\includegraphics[width=\textwidth]{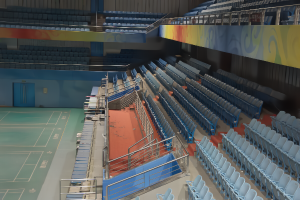}
		\caption{LLFLOW\cite{llflow}}
	\end{subfigure}
 %%%%%%%%%%%%%%
	\begin{subfigure}{.24\textwidth}
		\includegraphics[width=\textwidth]{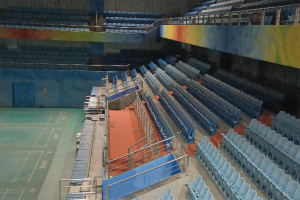}
		\caption{MIRNet-v2\cite{mirnetv2}}
	\end{subfigure}
  %%%%%%%%%%%%%%
	\begin{subfigure}{.24\textwidth}
		\includegraphics[width=\textwidth]{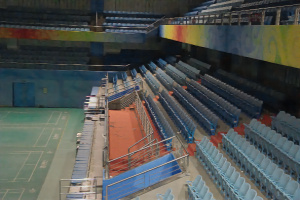}
		\caption{SNR-Aware\cite{snraware}}
	\end{subfigure}
 %%%%%%%%%%
 	\begin{subfigure}{.24\textwidth}
		\includegraphics[width=\textwidth]{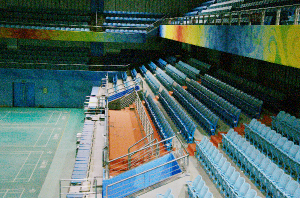}
		\caption{Zero-DCE\cite{zerodce}}
	\end{subfigure}
 %%%%%%%%%%%%%%
	\begin{subfigure}{.24\textwidth}
		\includegraphics[width=\textwidth]{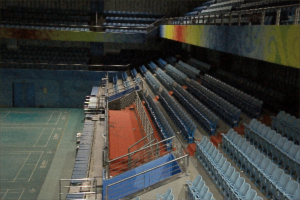}
		\caption{IAT\cite{90K}}
	\end{subfigure}
 %%%%%%%%%%%%%%
	\begin{subfigure}{.24\textwidth}
		\includegraphics[width=\textwidth]{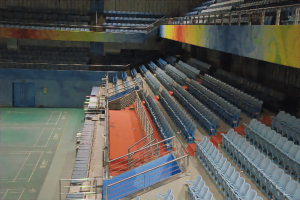}
		\caption{ours}
	\end{subfigure}
  %%%%%%%%%%%%%%
	\begin{subfigure}{.24\textwidth}
		\includegraphics[width=\textwidth]{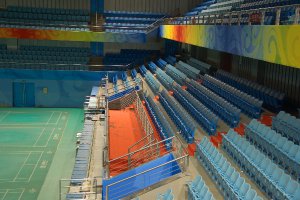}
		\caption{GT}
	\end{subfigure}
  %%%%%%%%%%%%%%
  %%%%%%%%%%%%%%
  	\begin{subfigure}{.24\textwidth}
		\includegraphics[width=\textwidth]{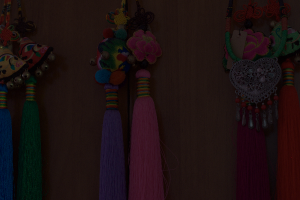}
		\caption{Input}
	\end{subfigure}
 %%%%%%%%%%%%%%
	\begin{subfigure}{.24\textwidth}
		\includegraphics[width=\textwidth]{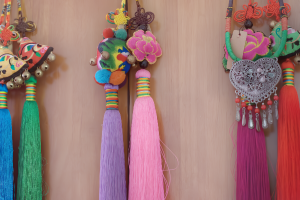}
		\caption{LLFLOW\cite{llflow}}
	\end{subfigure}
 %%%%%%%%%%%%%%
	\begin{subfigure}{.24\textwidth}
		\includegraphics[width=\textwidth]{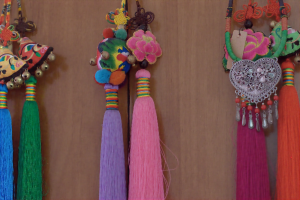}
		\caption{MIRNet-v2\cite{mirnetv2}}
	\end{subfigure}
  %%%%%%%%%%%%%%
	\begin{subfigure}{.24\textwidth}
		\includegraphics[width=\textwidth]{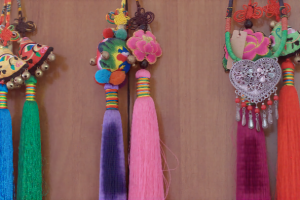}
		\caption{SNR-Aware\cite{snraware}}
	\end{subfigure}
 %%%%%%%%%%
 	\begin{subfigure}{.24\textwidth}
		\includegraphics[width=\textwidth]{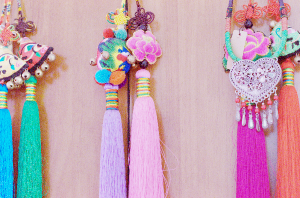}
		\caption{Zero-DCE \cite{zerodce}}
	\end{subfigure}
 %%%%%%%%%%%%%%
	\begin{subfigure}{.24\textwidth}
		\includegraphics[width=\textwidth]{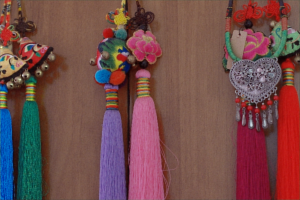}
		\caption{IAT \cite{90K}}
	\end{subfigure}
 %%%%%%%%%%%%%%
	\begin{subfigure}{.24\textwidth}
		\includegraphics[width=\textwidth]{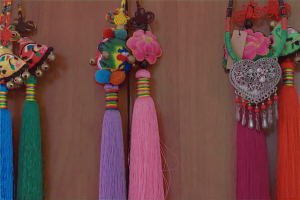}
		\caption{ours}
	\end{subfigure}
  %%%%%%%%%%%%%%
	\begin{subfigure}{.24\textwidth}
		\includegraphics[width=\textwidth]{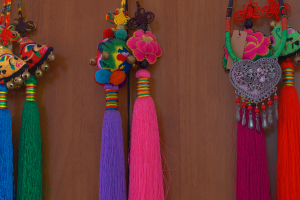}
		\caption{GT}
	\end{subfigure}

  %%%%%%%%%%%%%%

\caption{Visual comparison of our method on LOL-v1 dataset}
\label{fig:visualresults}
\end{figure*}

The loss function consists of two different components. Smooth L1 loss is preferred to ensure that the estimated image pixels are close to the normal-light images. The second loss is Multi-scale Structure Similarity Index Measure (MS-SSIM), which enforces the network to predict more visually pleasing image. The total loss function is calculated as:
\begin{align}
L_{TOTAL}=\alpha _{1}*L_{L1}+\alpha _{2}*L_{MS-SSIM}
\end{align}
$\alpha _{1}$ and $\alpha _{2}$ are hyperparameters for balancing the loss functions.

\subsection{Comparative LLIE Results}

%%%%%%%
State-of-the-art LLIE \cite{90K, snraware, llflow, kind,retinexnet,DRBN}  along with some image restoration networks \cite{maxim, mirnetv2} are selected for comparison. We report PSNR and SSIM results for LOL-v1, LOL-v2-real, LOL-v2-synthetic and Rellisur datasets in Tables \ref{tab:lolv1}, \ref{tab:lolv2r}, \ref{tab:lolv2s} and \ref{tab:rellisur} respectively and their corresponding comparative plots are given in Figure \ref{fig:performance}. 
\footnote{Note that in the literature, there are some methods \cite{llflow,kind} which are using some information coming from GT images while reporting performance values. However, for the sake of fairness, the results reported in this section assume that there is no GT information for any of the methods considered.} Our method achieve best PSNR results except the LOL-v2-real dataset. For SSIM metric, again, we achieve best performance in three of four datasets with the help of MS-SSIM loss.

Comparative visual results on two images from LOL-v1 dataset for qualitative analysis are presented in Figure \ref{fig:visualresults}. In the first image, the effect of our SDIA block can be observed in very dark regions. The output of second image proves the success of our color correction and denoising blocks. The color distribution for this image is the best by far compared to LLIE methods and better than Mirnet-v2 \cite{mirnetv2}. Also, our method does not produce any artifact like in the result of \cite{snraware} in third and fifth tassels from the left side.      

In order to show the effectiveness of the proposed solution in computation, we compare our method with the state-of-the-art techniques. We utilize a mobile workstation which has a GPU of NVIDIA GeForce RTX 3070M 8GB. The total parameters and inference times of our method and selected approaches are given in Table \ref{tab:computation}. Computation times are obtained by averaging over 100 runs for an image size of 600x400 which is the size of images in LOL-v1 dataset. As expected, our method is the fastest of all of the supervised networks. When PSNR values are compared with Zero-DCE \cite{zerodcepp}, the difference is larger than 10dB on LoL-v1 dataset. Another notable result is the computation time of the work in \cite{snraware}. Since it works on patches, its computation time is much lower than the other methods \cite{llflow} which has similar number of parameters.

%%%%%%%%
\begin{figure*}[ht!]
	\centering
	\begin{subfigure}{.23\textwidth}
		\includegraphics[width=\textwidth]{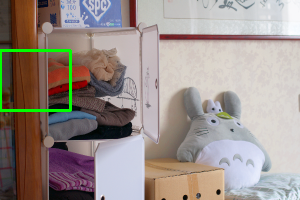}
		%\caption{GT}
	\end{subfigure}
 %%%%%%%%%%%%%%
    \begin{subfigure}{.18\textwidth}
		\includegraphics[width=\textwidth]{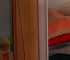}
		%\caption{SDIA}
	\end{subfigure}
 %%%%%%%%%%%%%%
	\begin{subfigure}{.18\textwidth}
		\includegraphics[width=\textwidth]{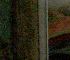}
		%\caption{GISP}
	\end{subfigure} 
  %%%%%%%%%%%%%%
	\begin{subfigure}{.18\textwidth}
		\includegraphics[width=\textwidth]{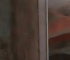}
		%\caption{FLIGHT-NET}
	\end{subfigure}
 %%%%%%%%%%%%%%
 	\begin{subfigure}{.18\textwidth}
		\includegraphics[width=\textwidth]{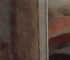}
		%\caption{FLIGHT-NET}
	\end{subfigure}
  %%%%%%%%%%%%%%
  %%%%%%%%%%%%%%
	\begin{subfigure}{.23\textwidth}
		\includegraphics[trim={0 0 42 0},clip,width=\textwidth]{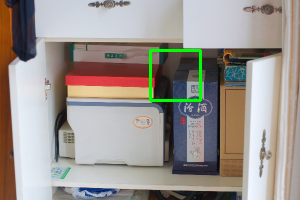}
		%\caption{GT}
	\end{subfigure}
 %%%%%%%%%%%%%%
    \begin{subfigure}{.18\textwidth}
		\includegraphics[width=\textwidth]{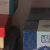}
		%\caption{SDIA}
	\end{subfigure}
 %%%%%%%%%%%%%%
	\begin{subfigure}{.18\textwidth}
		\includegraphics[width=\textwidth]{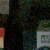}
		%\caption{GISP}
	\end{subfigure}
  %%%%%%%%%%%%%%
	\begin{subfigure}{.18\textwidth}
		\includegraphics[width=\textwidth]{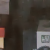}
		%\caption{FLIGHT-NET}
	\end{subfigure}
   %%%%%%%%%%%%%%
	\begin{subfigure}{.18\textwidth}
		\includegraphics[width=\textwidth]{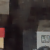}
		%\caption{FLIGHT-NET}
	\end{subfigure}
  %%%%%%%%%%%%%%
  %%%%%%%%%%%%%%
	\begin{subfigure}{.23\textwidth}
		\includegraphics[trim={0 35 0 20},clip,width=\textwidth]{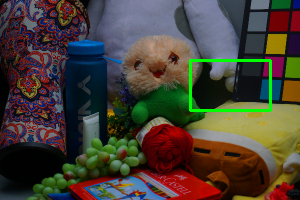}
		\caption{GT}
	\end{subfigure}
 %%%%%%%%%%%%%%
    \begin{subfigure}{.18\textwidth}
		\includegraphics[width=\textwidth]{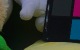}
		\caption{GT-Zoom}
	\end{subfigure}
 %%%%%%%%%%%%%%
	\begin{subfigure}{.18\textwidth}
		\includegraphics[width=\textwidth]{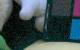}
		\caption{SDIA}
	\end{subfigure}
  %%%%%%%%%%%%%%
	\begin{subfigure}{.18\textwidth}
		\includegraphics[width=\textwidth]{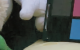}
		\caption{GISP}
	\end{subfigure}
   %%%%%%%%%%%%%%
	\begin{subfigure}{.18\textwidth}
		\includegraphics[width=\textwidth]{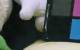}
		\caption{FLIGHT-NET}
	\end{subfigure}
  %%%%%%%%%%%%%%
\caption{Visual results of ablation study.}
\label{fig:ablation}
\end{figure*}

%%%%%%%

Although FLIGHT-Net has the lowest parameters with 25K compared to its main competitors \cite{90K, snraware, maxim}, it has the best PSNR value with 24.96 and second-best SSIM value with 0.85 on LOL-v1 dataset. The best result for SSIM belongs to MAXIM \cite{maxim} has 14.14M parameters which much more than the number of parameters of our method. IAT \cite{90K} can be considered as the main competitor of FLIGHT-Net when the number of parameters is taken into account since it is the only method with good performance among competitors with less than 100K parameters, however, its PSNR value is 23.38 which is much lower than the PSNR value of FLIGHT-Net. Zero-DCE has 10k parameters, but its performance is very far from state-of-art.

Depending on our observation, the reason of slightly lower PSNR values in LOL-v2-real dataset is the small misalignment between ground truth and training images. This fact is also the reason of not reporting the performance values of IAT \cite{90K} since it follows a different strategy during training to eliminate this misalignment for LOL-v2-real dataset. 

\subsection{Ablation Study}

In order to demonstrate the effectiveness of our subblocks and loss functions, five ablation studies are performed on LOL-v1 dataset. The quantitative results of the study are presented in Table \ref{table:ablation} and visual results are given in Figure \ref{fig:ablation}.

In the first stage of the ablation study, SDIA and GISP blocks are trained separately. As expected, the SDIA branch improved the illumination but also shows the necessity of GISP block. In output image obtained with SDIA alone, we have still spatial noise and color inconsistency and therefore low PSNR and SSIM values. The GISP module handles these issues as mentioned earlier. We also experience the performance of GISP module alone, however, its performance is far behind the overall proposed network. As expected, the SDIA branch increases visibility and transforms the image into the latent feature space required for GISP to fit the target image.

We also test the effects of smooth L1 and MS-SSIM loss functions. PSNR and SSIM values for different combinations of loss functions are reported in Table \ref{table:ablation}. Training with only smooth L1 or MS-SSIM loss functions are not enough to get the optimum results. PSNR values in the case of using smooth L1 and MS-SSIM loss functions are 22.51 and 22.94 respectively while PSNR value is 24.96 in the case of using both loss functions. As a result, FLIGHT-Net achieves state-of-the-art performance when trained with the combination of smooth L1 and MS-SSIM loss.
%%%%%%%%

%%%%%%%

%%%
\begin{table}[htp]
\centering
\begin{adjustbox}{width=\textwidth/2}

\begin{tabular}{|l|c|c|l|l|}

\hline
Blocks    & Smooth L1 & MS-SSIM & PSNR & SSIM  \\ \hline
SDIA      & \checkmark              & \checkmark                              & 20.78                    & 0.73  \\
GISP      & \checkmark              & \checkmark                              & 23.70                     & 0.83  \\
SDIA+GISP & \checkmark              & \XSolid                              & 22.51                    & 0.80  \\
SDIA+GISP & \XSolid              & \checkmark                              & 22.94                    & 0.84  \\
SDIA+GISP & \checkmark              & \checkmark                              & 24.96                   & 0.85 \\ \hline
\end{tabular}
\end{adjustbox}
\caption{Results of the ablation study on LOL-v1 dataset}
\vspace{-16pt}
\label{table:ablation}
\end{table}

%%%%%%%%%%%%
\section{Conclusion}
In this paper, we have proposed a lightweight yet effective framework for low-light image enhancement problem. Our solution has achieved a state-of-the-art performance for the problem on several datasets with one of the lightest network in the literature. We believe that carefully investigating the forward problem formulation and image signal processing framework and designing the blocks accordingly helps reducing the number of parameters and boosting performance rather than hoping a pure huge bulky DNN to extract and solve for the hidden relations among the features. Lastly, as a future work, we have plans to extend our approach for the processing of low-light videos by considering temporally and spatially varying light conditions.

%%%%%%%%% REFERENCES
{\small
\bibliographystyle{ieee_fullname}
\bibliography{main}

\begin{thebibliography}{10}\itemsep=-1pt

\bibitem{rellisur_aakerberg2021}
Andreas Aakerberg, Kamal Nasrollahi, and Thomas~B Moeslund.
\newblock Rellisur: A real low-light image super-resolution dataset.
\newblock In {\em Thirty-fifth Conference on Neural Information Processing
  Systems-NeurIPS 2021}, 2021.

\bibitem{IPT_chen2021pre}
Hanting Chen, Yunhe Wang, Tianyu Guo, Chang Xu, Yiping Deng, Zhenhua Liu, Siwei
  Ma, Chunjing Xu, Chao Xu, and Wen Gao.
\newblock Pre-trained image processing transformer.
\newblock In {\em Proceedings of the IEEE/CVF Conference on Computer Vision and
  Pattern Recognition}, pages 12299--12310, 2021.

\bibitem{90K}
Ziteng Cui, Kunchang Li, Lin Gu, Shenghan Su, Peng Gao, Zhengkai Jiang, Yu
  Qiao, and Tatsuya Harada.
\newblock You only need 90k parameters to adapt light: A light weight
  transformer for image enhancement and exposure correction.
\newblock In {\em BMVC}, 2022.

\bibitem{dipbook}
Rafael~C Gonzalez.
\newblock {\em Digital image processing}.
\newblock Pearson Education, 2009.

\bibitem{zerodce}
Chunle Guo, Chongyi Li, Jichang Guo, Chen~Change Loy, Junhui Hou, Sam Kwong,
  and Runmin Cong.
\newblock Zero-reference deep curve estimation for low-light image enhancement.
\newblock In {\em Proceedings of the IEEE/CVF conference on computer vision and
  pattern recognition}, pages 1780--1789, 2020.

\bibitem{lime}
Xiaojie Guo, Yu Li, and Haibin Ling.
\newblock Lime: Low-light image enhancement via illumination map estimation.
\newblock {\em IEEE Transactions on image processing}, 26(2):982--993, 2016.

\bibitem{channelatt}
Jie Hu, Li Shen, and Gang Sun.
\newblock Squeeze-and-excitation networks.
\newblock In {\em Proceedings of the IEEE conference on computer vision and
  pattern recognition}, pages 7132--7141, 2018.

\bibitem{uw1}
Zhixiong Huang, Jinjiang Li, Zhen Hua, and Linwei Fan.
\newblock Underwater image enhancement via adaptive group attention-based
  multiscale cascade transformer.
\newblock {\em IEEE Transactions on Instrumentation and Measurement}, 71:1--18,
  2022.

\bibitem{ignatov2020replacing}
Andrey Ignatov, Luc Van~Gool, and Radu Timofte.
\newblock Replacing mobile camera isp with a single deep learning model.
\newblock In {\em Proceedings of the IEEE/CVF Conference on Computer Vision and
  Pattern Recognition Workshops}, pages 536--537, 2020.

\bibitem{uw2}
Zhiying Jiang, Zhuoxiao Li, Shuzhou Yang, Xin Fan, and Risheng Liu.
\newblock Target oriented perceptual adversarial fusion network for underwater
  image enhancement.
\newblock {\em IEEE Transactions on Circuits and Systems for Video Technology},
  32(10):6584--6598, 2022.

\bibitem{retinex}
Edwin~H Land.
\newblock An alternative technique for the computation of the designator in the
  retinex theory of color vision.
\newblock {\em Proceedings of the national academy of sciences},
  83(10):3078--3080, 1986.

\bibitem{zerodcepp}
Chongyi Li, Chunle~Guo Guo, and Chen~Change Loy.
\newblock Learning to enhance low-light image via zero-reference deep curve
  estimation.
\newblock In {\em IEEE Transactions on Pattern Analysis and Machine
  Intelligence}, 2021.

\bibitem{LP_Net_li2020luminance}
Jiaqian Li, Juncheng Li, Faming Fang, Fang Li, and Guixu Zhang.
\newblock Luminance-aware pyramid network for low-light image enhancement.
\newblock {\em IEEE Transactions on Multimedia}, 23:3153--3165, 2020.

\bibitem{dslr}
Seokjae Lim and Wonjun Kim.
\newblock Dslr: Deep stacked laplacian restorer for low-light image
  enhancement.
\newblock {\em IEEE Transactions on Multimedia}, 23:4272--4284, 2020.

\bibitem{ruas}
Risheng Liu, Long Ma, Jiaao Zhang, Xin Fan, and Zhongxuan Luo.
\newblock Retinex-inspired unrolling with cooperative prior architecture search
  for low-light image enhancement.
\newblock In {\em Proceedings of the IEEE/CVF Conference on Computer Vision and
  Pattern Recognition}, pages 10561--10570, 2021.

\bibitem{liu2022convnet}
Zhuang Liu, Hanzi Mao, Chao-Yuan Wu, Christoph Feichtenhofer, Trevor Darrell,
  and Saining Xie.
\newblock A convnet for the 2020s.
\newblock In {\em Proceedings of the IEEE/CVF Conference on Computer Vision and
  Pattern Recognition}, pages 11976--11986, 2022.

\bibitem{llnet}
Kin~Gwn Lore, Adedotun Akintayo, and Soumik Sarkar.
\newblock Llnet: A deep autoencoder approach to natural low-light image
  enhancement.
\newblock {\em Pattern Recognition}, 61:650--662, 2017.

\bibitem{mbllen_lv2018}
Feifan Lv, Feng Lu, Jianhua Wu, and Chongsoon Lim.
\newblock Mbllen: Low-light image/video enhancement using cnns.
\newblock In {\em BMVC}, volume 220, page~4, 2018.

\bibitem{ispeccv22eth}
Ardhendu Shekhar~Tripathi, Martin Danelljan, Samarth Shukla, Radu Timofte, and
  Luc Van~Gool.
\newblock Transform your smartphone into a dslr camera: Learning the isp in the
  wild.
\newblock In {\em Computer Vision--ECCV 2022: 17th European Conference, Tel
  Aviv, Israel, October 23--27, 2022, Proceedings, Part VI}, pages 625--641.
  Springer, 2022.

\bibitem{maxim}
Zhengzhong Tu, Hossein Talebi, Han Zhang, Feng Yang, Peyman Milanfar, Alan
  Bovik, and Yinxiao Li.
\newblock Maxim: Multi-axis mlp for image processing.
\newblock In {\em Proceedings of the IEEE/CVF Conference on Computer Vision and
  Pattern Recognition}, pages 5769--5780, 2022.

\bibitem{gladnet_wang2018}
Wenjing Wang, Chen Wei, Wenhan Yang, and Jiaying Liu.
\newblock Gladnet: Low-light enhancement network with global awareness.
\newblock In {\em 2018 13th IEEE international conference on automatic face \&
  gesture recognition (FG 2018)}, pages 751--755. IEEE, 2018.

\bibitem{llflow}
Yufei Wang, Renjie Wan, Wenhan Yang, Haoliang Li, Lap-Pui Chau, and Alex Kot.
\newblock Low-light image enhancement with normalizing flow.
\newblock In {\em Proceedings of the AAAI Conference on Artificial
  Intelligence}, volume~36, pages 2604--2612, 2022.

\bibitem{retinexnet}
Chen Wei, Wenjing Wang, Wenhan Yang, and Jiaying Liu.
\newblock Deep retinex decomposition for low-light enhancement.
\newblock In {\em BMVC}, 2018.

\bibitem{Retinex-Net_wei2018deep}
Chen Wei, Wenjing Wang, Wenhan Yang, and Jiaying Liu.
\newblock Deep retinex decomposition for low-light enhancement.
\newblock {\em arXiv preprint arXiv:1808.04560}, 2018.

\bibitem{cbam}
Sanghyun Woo, Jongchan Park, Joon-Young Lee, and In~So Kweon.
\newblock Cbam: Convolutional block attention module.
\newblock In {\em Proceedings of the European conference on computer vision
  (ECCV)}, pages 3--19, 2018.

\bibitem{snraware}
Xiaogang Xu, Ruixing Wang, Chi-Wing Fu, and Jiaya Jia.
\newblock Snr-aware low-light image enhancement.
\newblock In {\em Proceedings of the IEEE/CVF Conference on Computer Vision and
  Pattern Recognition}, pages 17714--17724, 2022.

\bibitem{DRBN}
Wenhan Yang, Shiqi Wang, Yuming Fang, Yue Wang, and Jiaying Liu.
\newblock From fidelity to perceptual quality: A semi-supervised approach for
  low-light image enhancement.
\newblock In {\em Proceedings of the IEEE/CVF conference on computer vision and
  pattern recognition}, pages 3063--3072, 2020.

\bibitem{band_yang2021}
Wenhan Yang, Shiqi Wang, Yuming Fang, Yue Wang, and Jiaying Liu.
\newblock Band representation-based semi-supervised low-light image
  enhancement: Bridging the gap between signal fidelity and perceptual quality.
\newblock {\em IEEE Transactions on Image Processing}, 30:3461--3473, 2021.

\bibitem{sparseTIP2021}
Wenhan Yang, Wenjing Wang, Haofeng Huang, Shiqi Wang, and Jiaying Liu.
\newblock Sparse gradient regularized deep retinex network for robust low-light
  image enhancement.
\newblock {\em IEEE Transactions on Image Processing}, 30:2072--2086, 2021.

\bibitem{cycleISP}
Syed~Waqas Zamir, Aditya Arora, Salman Khan, Munawar Hayat, Fahad~Shahbaz Khan,
  Ming-Hsuan Yang, and Ling Shao.
\newblock Cycleisp: Real image restoration via improved data synthesis.
\newblock In {\em CVPR}, pages 2696--2705, 2020.

\bibitem{mirnetv1}
Syed~Waqas Zamir, Aditya Arora, Salman Khan, Munawar Hayat, Fahad~Shahbaz Khan,
  Ming-Hsuan Yang, and Ling Shao.
\newblock Learning enriched features for real image restoration and
  enhancement.
\newblock In {\em Computer Vision--ECCV 2020: 16th European Conference,
  Glasgow, UK, August 23--28, 2020, Proceedings, Part XXV 16}, pages 492--511.
  Springer, 2020.

\bibitem{mirnetv2}
Syed~Waqas Zamir, Aditya Arora, Salman Khan, Munawar Hayat, Fahad~Shahbaz Khan,
  Ming-Hsuan Yang, and Ling Shao.
\newblock Learning enriched features for fast image restoration and
  enhancement.
\newblock {\em IEEE Transactions on Pattern Analysis and Machine Intelligence},
  45(2):1934--1948, 2022.

\bibitem{CA_zhang2018image}
Yulun Zhang, Kunpeng Li, Kai Li, Lichen Wang, Bineng Zhong, and Yun Fu.
\newblock Image super-resolution using very deep residual channel attention
  networks.
\newblock In {\em Proceedings of the European conference on computer vision
  (ECCV)}, pages 286--301, 2018.

\bibitem{kind}
Yonghua Zhang, Jiawan Zhang, and Xiaojie Guo.
\newblock Kindling the darkness: A practical low-light image enhancer.
\newblock In {\em Proceedings of the 27th ACM international conference on
  multimedia}, pages 1632--1640, 2019.

\bibitem{retinexdip}
Zunjin Zhao, Bangshu Xiong, Lei Wang, Qiaofeng Ou, Lei Yu, and Fa Kuang.
\newblock Retinexdip: A unified deep framework for low-light image enhancement.
\newblock {\em IEEE Transactions on Circuits and Systems for Video Technology},
  32(3):1076--1088, 2021.

\bibitem{clahe}
Karel Zuiderveld.
\newblock Contrast limited adaptive histogram equalization.
\newblock {\em Graphics gems}, pages 474--485, 1994.

\end{thebibliography}
}

\end{document}